\pdfoutput=1

\documentclass[11pt]{article}

\usepackage{acl}

\usepackage{times}
\usepackage{latexsym}
\usepackage{float}
\usepackage{placeins}

\usepackage[T1]{fontenc}

\usepackage[utf8]{inputenc}

\usepackage{microtype}

\usepackage{inconsolata}

\usepackage{natbib}
\usepackage{multibib}
\usepackage{amssymb}
\usepackage{amsmath}
\usepackage{graphicx}
\usepackage{tabularx}
\usepackage{soul}
\usepackage{multirow}
\usepackage{booktabs}
\usepackage{enumitem}

\usepackage{xcolor}
\usepackage{hyperref}
 \definecolor{darkblue}{rgb}{0, 0, 0.5}
  \hypersetup{colorlinks=true, citecolor=darkblue, linkcolor=darkblue, urlcolor=darkblue}

\usepackage{xstring}

\usepackage{color}
\usepackage{mathrsfs}  

\title{GOLD: Generalized Knowledge Distillation via Out-of-Distribution-Guided Language Data Generation}

\author{Mohsen Gholami\textsuperscript{\rm 1,2}, Mohammad Akbari\textsuperscript{\rm 1}, Cindy Hu\textsuperscript{\rm 1}, \bf Vaden Masrani\textsuperscript{\rm 1}, 
\\
{\bf Z. Jane Wang\textsuperscript{\rm 2}}, \and {\bf Yong Zhang\textsuperscript{\rm 1}} \\
        \textsuperscript{\rm 1}Huawei Technologies Canada Co. Ltd.\\ \textsuperscript{\rm 2}University of Biritish Columbia \\ \texttt{\{mohsen.gholami, mohammad.akbari, cindy.hu1, yong.zhang3\}@huawei.com}}

\begin{document}
\maketitle
\begin{abstract}
Knowledge distillation from LLMs is essential for the efficient deployment of language models. Prior works have proposed data generation using LLMs for preparing distilled models. We argue that generating data with LLMs is prone to sampling mainly from the center 
of original content distribution. This limitation hinders the distilled model from learning the true underlying data distribution and to forget the tails of the distributions (samples with lower probability). To this end, we propose \emph{GOLD} 
, a task-agnostic data generation and knowledge distillation framework, which employs an iterative out-of-distribution-guided feedback mechanism for the LLM. As a result, the generated data improves the generalizability of distilled models. An energy-based OOD evaluation approach is also introduced to deal with noisy generated data. Our extensive experiments on 10 different classification and sequence-to-sequence tasks in NLP show that \emph{GOLD} respectively outperforms prior arts and the LLM with an average improvement of 5\% and 14\%. We will also show that the proposed method is applicable to less explored and novel tasks. {Code is available} 
\href{https://developer.huaweicloud.com/develop/aigallery/notebook/detail?id=9d770d1f-3758-4d0f-99d4-3346abbe1546}
{here\footnote{\url{https://developer.huaweicloud.com/develop/aigallery/notebook/detail?id=9d770d1f-3758-4d0f-99d4-3346abbe1546}}}.
\end{abstract}

\section{Introduction}
Large language models (LLMs) have shown outstanding few-shot performance in solving different complex natural language tasks \cite{brown2020language}. 
{The term few-shot refers to the ability of the LLM to understand and perform tasks accurately given only a few examples.}
However, achieving such performance necessitates models with a large number of parameters to generalize and learn distinct tasks. The computational complexity of LLMs hinders their real-world applications for deployment. 
Moreover, most of the LLMs are not publicly available and users should share their confidential data with LLMs through prompting which is a privacy concern. 
Therefore, user-specific small models are critical to address efficiency and privacy concerns.  

To circumvent the above challenges, knowledge distillation (KD) from LLMs has been used to prepare small language models (SLMs). 
There are two main paradigms in KD: \textit{data-informed} and \textit{data-free} methods \cite{GKD, step-by-step, minillm}. Data-informed methods are the conventional KD techniques that use LLMs to label data and train a small language model with the labels \cite{GKD, step-by-step}. 
Data-free methods, on the other hand, study an extreme case when no human-generated dataset (unlabeled or labeled) is available during distillation. ZeroGen \cite{zerogen} and ProGen \cite{progen} are two prior works that proposed data-free knowledge distillation.

\begin{figure}[t]
    \centering
    \includegraphics[scale=0.50]{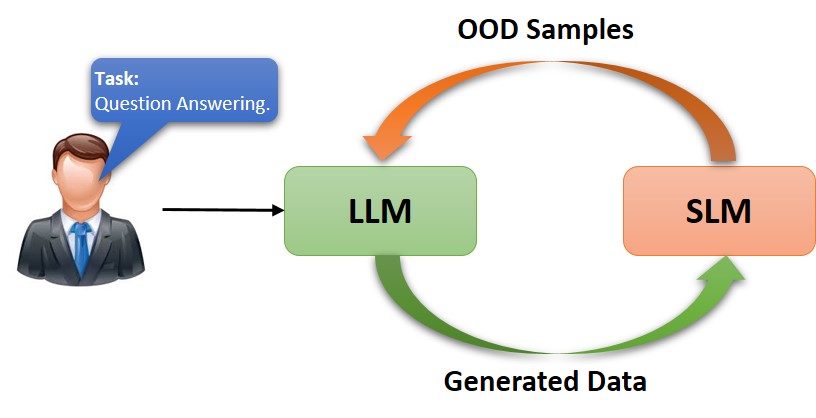}
    \vspace{-5pt}
    \caption{\emph{GOLD} finds failure modes of SLM in the course of data generation and guides the LLM to generate OOD samples to improve SLM's generalizability.
    }
    \label{fig:tiser}
    \vspace{5pt}
\end{figure}

ZeroGen \cite{zerogen} proposed generating data for classification tasks using LLMs and trained a small model on the generated data. ProGen \cite{progen} improved the quality of the generated data by finding important samples via an influence function in the course of data generation. 
These works are not able to find failure modes of SLMs and steer data generation toward the samples that can improve the generalizability of the model.

\citep{curseofrecursion} showed that distilling knowledge from LLMs via data generation causes irreversible defects in the SLM, where the tails of the original content distribution disappear. LLMs usually tend to generate samples with higher likelihood (known as in-distribution samples) repeatedly \citep{curseofrecursion} which results in poor generalizability of the distilled SLM. Preserving the LLM's ability to model low-probability events {or out-of-distribution (OOD) samples} is essential to the fairness of their predictions.
Such events are often relevant to marginalized groups and are vital to understand complex systems \cite{21822c09-6ede-348c-b91e-4e92777f2281}.

Usually, the performance of language models improves when more human-labeled data is available, since a larger dataset covers wider task domains. On the contrary, \citep{falsepromise} shows that increasing the amount of generated data with LLMs lowers the performance of the distilled SLM. 
We argue that vanilla data generation with LLMs is prone to only generate high likelihood samples which negatively affects the SLM performance with increasing the number of samples. 

\begin{figure*}[t]
    \centering
    \includegraphics[scale=0.35]{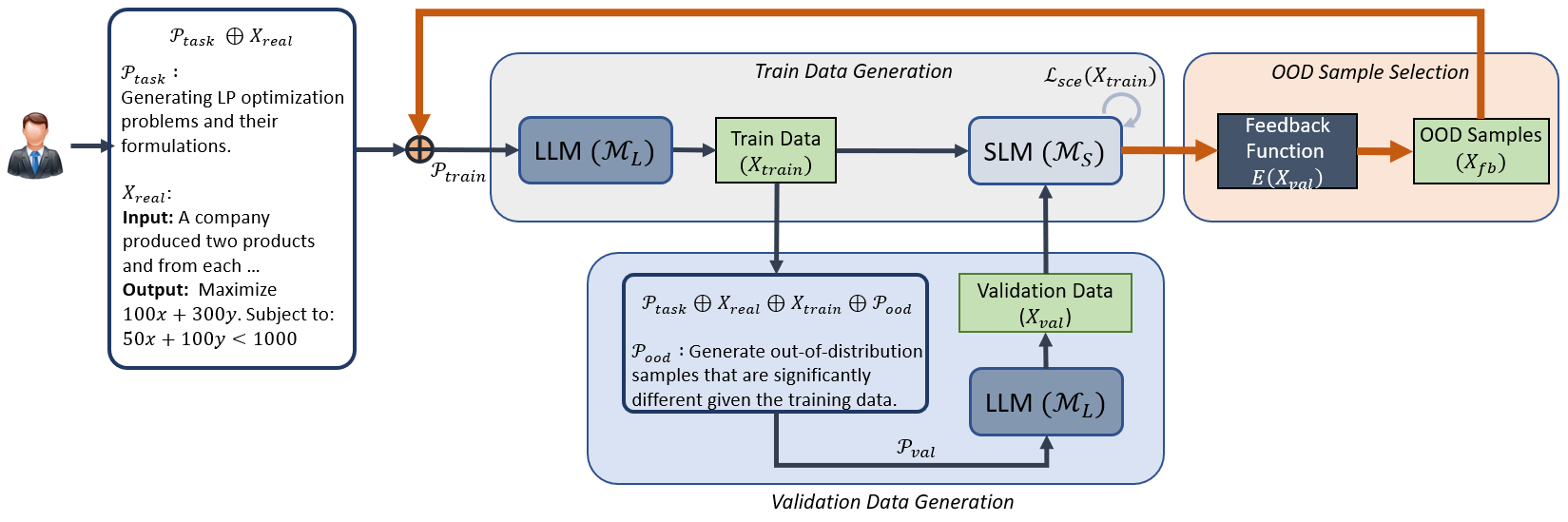}
    \vspace{-5pt}
    \caption{Overview of the proposed data generation and knowledge distillation method, \emph{GOLD}. $\oplus$: Concatenation.}
    \label{fig:framework}
\end{figure*}

To this end, we propose \emph{GOLD}, a {data-free} KD framework that {iteratively} finds failure modes of the SLM (i.e., OOD samples) and provides feedback to the LLM (Figure \ref{fig:tiser}) for the next iteration of data generation. Figure \ref{fig:framework} shows the overall framework of \emph{GOLD}. The user provides the task definition with a few samples of the data corresponding to that task. The LLM is then used to generate a training batch of data for the specified task, which is used to update the weights of the SLM. Given the generated train batch, we use prompting to ask LLM to generate a separate batch of OOD data that is significantly different from the train batch in terms of topic and style 
This OOD batch is then used as our validation set to evaluate the performance of the SLM and accordingly identify its failure modes as a feedback to the LLM.
We use the output logits of the SLM to measure the energy values of the validation samples and select the top ones based on their corresponding free energy scores \cite{energy,akbari2023system}. The energy score does not require the labels of the generated sample and therefore is not prone to pick the data with noisy labels as the OOD samples. The selected OOD samples are then used as the feedback to the LLM for the next iteration of data generation. 
The major contributions of this paper are:
\begin{itemize}[leftmargin=*]
    \item {Proposing a task-agnostic framework for data generation and KD from LLMs to SLMs that is applicable to any NLP task, even novel tasks.}
    \item {Introducing an iterative OOD-empowered feedback mechanism to improve the generalizability of distilled SLMs.}
    \item {Proposing an energy-based OOD evaluation approach to handle noisy data generated by LLMs.}
    \item {Achieving state-of-the-art results on a variety of classification and sequence-to-sequence downstream tasks in NLP.}
\end{itemize}


\vspace{-5pt}
\section{Related Works}

\textbf{Data-Informed KD.} Distil-Step-by-Step \citep{step-by-step} is a data-informed approach which proposes to perform KD from an LLM as an annotation function for real data, where the annotations take the form of ``rationals'' extracted via Chain-of-Thought prompting. 
Other data-informed methods such as MiniLLM \citep{minillm} and GKD \citep{GKD} also propose to change the training objective used for KD away from the commonly-used forward Kullback-Leibler divergence between the teacher and student distribution. 

\textbf{Data-Free KD.} ZeroGen \citep{zerogen} and ProGen \citep{progen} are two data-free approaches to train task-specific student models with much fewer parameters than the base model. Using task-specific prompting in place of human annotations, and a sufficiently large generated train set, \citep{zerogen} can \textit{exceed} the performance of the base model on text classification, question answering, and natural language inference. ProGen \citep{progen} then builds on ZeroGen by using an ``influence function'' to measure the quality of the generated samples on a synthetic validation set, and uses this to reduce the portion of synthetic data which is low-quality or redundant. ProGen achieves on-par or superior performance with only 1\% synthetic dataset size compared to ZeroGen.


\textbf{Limitations of Data Generation with LLMs.} Two recent works \citep{falsepromise,curseofrecursion} challenge the idea that training a generative model 
on the output of a different generative model
will lead to performance gains in the case of a smaller ``imitation models'' being trained to match the performance of a ``teacher model'' as in \citep{falsepromise}.
However, we note their negative findings only apply in the most general case, where the synthetic train data is not ``corrected'' after being generated, either algorithmically or manually by humans before being hosted onto the web. Further, the future-generation models considered in both works are not ``task-specific'' - they are trained with the goal of either meeting or exceeding the performance of the previous model, and the negative findings apply only in this case. Our goals are less ambitious, and tailored towards achieving the performance of the teacher model (i.e., an LLM) only on a \textit{specific} task, rather than training a student model (i.e., an SLM) to reproduce the entire suite of capabilities as the LLM. 

{In addition, \citep{curseofrecursion} indicates the poor generalizability of the distilled SLM due to the generation of only high likelihood data by the LLM. However, unlike the previous works, \emph{GOLD} can effectively address this issue by iterative OOD-based feedbacks to the LLM to identify the failure modes of the SLM and improve its generalizability.}

\vspace{-3pt}
\section{Method}
\vspace{-4pt}

Fig. \ref{fig:framework} shows an overview of the proposed \emph{GOLD}, a {data generation and} KD framework for language models. We iteratively transfer knowledge from an LLM, denoted by $\mathcal{M}_{L}$, to an SLM, denoted by $\mathcal{M}_{S}$, via the data generated by $\mathcal{M}_{L}$. The KD is performed for any task that falls within the realm of the expertise of $\mathcal{M}_{L}$. The framework operates with two inputs: a task definition $\mathcal{P}_{task}$ and a few samples of real data $X_{\text{real}}$. 

In each iteration, \emph{GOLD} generates a train set and an OOD validation set denoted by ($X_{\text{train}}, X_{\text{val}}$) that is respectively used to update the weights of $\mathcal{M}_S$ and to evaluate the updated $\mathcal{M}_S$. 
{A feedback function, denoted by $E$, is then used to find the top OOD samples from $X_{\text{val}}$ 
that are within a pre-defined upper and lower bound threshold. The sample selection procedure is performed 
regardless of the generated labels, which makes the feedback function robust to noisy labels.}
The selected samples, denoted by $X_{\text{fb}}$, are concatenated to the prompt 
for the next iteration of data generation. In the following, the details of the problem formulation, data generation procedure, the feedback function, and SLM training are discussed.

\subsection{Problem Formulation}
Our objective is to enhance the train data $X_{\text{train}}$ to address failure modes of $\mathcal{M}_S$. 
This is achieved by automatically selecting the in-context examples present in the prompt {$\mathcal{P}_{train}$} (Fig. \ref{fig:framework}) such that when $\mathcal{M}_{S}$ is trained on $X_{\text{train}}$, it exhibits optimal performance on $X_{\text{val}}$. 
$X_{\text{val}}$ is generated to be \textit{significantly different} from $X_{\text{train}}$ in terms of topic and style in order to maximize the error of the SLM $\mathcal{M}_{S}$ on $X_{\text{val}}$. 
We argue that optimal knowledge transfer from $\mathcal{M}_{L}$  to $\mathcal{M}_{S}$ has been achieved when $\mathcal{M}_{L}$ only generates samples which $\mathcal{M}_{S}$ has already learned.
To facilitate this, it is necessary to push $\mathcal{M}_{L}$ to generate less likely (i.e., away from the modes of its training distribution), yet accurate samples.
It is important to note that LLMs often tend to generate high-likelihood samples, which can result in repeated or thematically similar samples.

We introduce a formulation that provides a conceptual framework for understanding our general approach and its underlying principles. Let denote the error of $\mathcal{M}_{S}$ on $X_{\text{val}}$ as $\mathcal{E}\big(\mathcal{M}_{S}( X_{\text{val}}|X_{\text{train}})\big)$, the objective function can be written as:
\vspace{-2pt}
\begin{equation}
\min_{X_{\text{train}}} \max_{X_{\text{val}}} \mathcal{E}\big(\mathcal{M}_{S}( X_{\text{val}}|X_{\text{train}})\big).
\vspace{-2pt}
\end{equation}

Unlike conventional optimization, the optimization of $\mathcal{E}$ leverages in-context learning where parameters of $\mathcal{M}_L$ are fixed. During each iteration, in-context examples in the prompt {$\mathcal{P}_{\text{train}}$} guide $X_{\text{train}}$ towards the failure modes of $\mathcal{M}_S$ (refer to Fig. \ref{fig:framework}). Similarly, {$X_{\text{val}}$} is steered towards more challenging samples using in-context examples in the prompt $\mathcal{P}_{\text{val}}$ (also see Fig. \ref{fig:framework}).

\subsection{Data Generation}
In the first iteration, given the prompt $\mathcal{P}_{train}$ including: 1) the task definition $\mathcal{P}_{\text{task}}$, and 2) $n$ human-labeled data samples (with their labels), denoted by $X_{\text{real}}$,
$\mathcal{M}_L$ generates a {train} batch of data:
\begin{equation}
    X_{\text{train}}^{t_0}=\mathcal{M}_L(\mathcal{P}_{\text{task}} \oplus X_{\text{real}}),
\end{equation}
where $\oplus$ denotes concatenation. $\mathcal{M}_L$ samples $X_{\text{train}}^{t_0}$ from the high likelihood tokens given the input prompt. Data generation with the aforementioned fixed prompt typically results in a distribution characterized by a high-density center. 
As indicated in prior works, such data generation procedure often leads to the disappearance of the tails in the distribution of human data observed by $\mathcal{M}_L$ \cite{curseofrecursion}. To address this issue, we initially propose the generation of an OOD validation set (with their labels). 


At this stage, the OOD validation samples are defined in relation to $X_{\text{train}}^{t_0}$, which are considered as the in-distribution samples. Thus, the validation data at iteration $t$ is generated by the prompt $\mathcal{P}_{\text{val}}=\mathcal{P}_{\text{task}} \oplus X_{\text{real}} \oplus X^{t}_{\text{train}} \oplus \mathcal{P}_{\text{ood}}$ as follows:
\begin{equation}
    X^{t}_{\text{val}}=\mathcal{M}_L(\mathcal{P}_{\text{task}} \oplus X_{\text{real}} \oplus X^{t}_{\text{train}} \oplus \mathcal{P}_{\text{ood}}),
\end{equation}
where $\mathcal{P}_{\text{ood}}$ prompts the generation of new data that significantly diverges from $X_{\text{train}}^{t}$, potentially in terms of topic, domain, or style.
$\mathcal{M}_S$ is evaluated on $X^{t}_{\text{val}}$ and based on the output logits, we identify failure samples (OOD samples) and form $X^{t}_{\text{fb}}$.
Detailed selection procedure of $X^{t}_{\text{fb}}$ is discussed in the next section.

The train data in the next iterations is generated by leveraging the feedback from the previous round. Given the prompt $\mathcal{P}_{train}$ consisting of task definition as well as real data and feedback samples, the train data at iteration $t+1$ is generated as follows:
\begin{equation}
    X_{\text{train}}^{t\text{+}1}=\mathcal{M}_L(\mathcal{P}_{\text{task}} \oplus X_{\text{real}} \oplus X^{t}_{\text{fb}}).
\end{equation}
Ideally, the train samples $X_{\text{train}}^{t\text{+}1}$ should be accurate and closely resemble the provided feedback samples {$X^{t}_{\text{fb}}$}. By updating the weights of the SLM $\mathcal{M}_S$ with these newly generated samples $X_{\text{train}}^{t+1}$, we can address the failure modes of $\mathcal{M}_S$ encountered in the previous iteration and enhance its generalizability. In other words, the process of validation and train data generation acts as two competing agents. Their goal is to improve the generalizability of the SLM and to challenge it with a progressively harder validation set.

\begin{table*}[!h]
\footnotesize
    \centering
    \begin{tabular}{p{1cm}|p{1.2cm}|p{12.3cm}} 
    \toprule
        {Task}&{Method}&Prompt \\ \midrule
        \multirow{2}{*}{ANLI}& \emph{\textbf{GOLD}}, ZeroGen, ProGen, P2Model&\textcolor{blue}{[Below are 3 examples of an natural language inference dataset. Samples include a ``hypothesis'' and a ``premise''. The label of the sample is: 1. ``entailment'' if the premise entails the hypothesis, 2. ``neutral' if the premise neither entails nor contradicts the hypothesis. 3. ``contradiction'' if premise contradicts hypothesis.]$^{\mathcal{P}1}$}  \textcolor{olive}{[Sample 1, Sample 2, Sample 3]$^{\mathcal{P}2}$} \textcolor{green}{[Generate a novel sample of data.]}  \\ \cline{2-3}
       &  Few-shot& \textcolor{blue}{[...]$^{\mathcal{P}1}$}\textcolor{olive}{[...]$^{\mathcal{P}2}$}\textcolor{darkblue}{[What is the label of the below sample?] [some sample]}\\ \bottomrule
    \end{tabular}
    \vspace{-3pt}
    \caption{Example prompts used for data generation. 
    \textit{Few-shot}: inference with the LLM given a few data samples. 
 }
    \label{tab:prompts}
\end{table*}

\subsection{OOD-Based Feedback}

\emph{GOLD} employs an energy function to detect OOD samples, which are then used as a feedback to the LLM. Energy functions have shown to be effective in identifying OOD samples in classification and regression tasks \citep{energy,ebjr,etran}. The concept of energy-based models (EBM) is pivotal in this context. 

Considering $\mathcal{M}_S$ as a classification model with $C$ classes, we can compute the free energy {score} corresponding to an input sample $x$
as follows:
\vspace{-6pt}
\begin{equation}
\vspace{-6pt}
E(x)=-\log\sum_c^C e^{\mathcal{M}_S^{(c)}(x)},
\label{eq:energy_f3}
\end{equation}
where $\mathcal{M}_S^{(c)}(x)$ is the output logit of $c$-th class.

{The above-mentioned procedure can also be extended to sequence-to-sequence models \citep{elang}, where a sequence of tokens is generated as output. Given $\mathcal{M}_S$ as our model, }
for each token, $\mathcal{M}_S$ predicts a class from a dictionary consisting of $K$ vocabularies. 
Consider an input-output $(x,\hat{y},l)$, where $x$ is the input text, $\hat{y}$ is the predicted sequence text, and $l$ is the corresponding logits of $\hat{y}$, denoted as $l:[{l}_i]_{i=1}^L$ where $L$ is the length of the returned sequence and ${l}_i \in \mathbb{R}^{K}$. Following Eq. \ref{eq:energy_f3}, we calculate the energy score of the input sequence $x$ as:
\vspace{-7pt}
\begin{equation}
E_s(x)=-\frac{1}{L}\sum_{i=1}^{L}\log\sum_{k=1}^K e^{l^{k}_{i}}
\label{eq:energy_f4},
\vspace{-3pt}
\end{equation}
that returns the average energy score over all tokens in the predicted sequence. The energy score calculation is performed in an unsupervised manner, where the data labels are not required.
Thus, we expect the energy score to be robust to potential noisy labels generated by the LLM.
We then select the samples with low negative energy scores from $X_{\text{val}}$ as follows:
\vspace{-3pt}
\begin{align}
    X_\text{fb} = \{X^{i}_{\text{val}} \mid  \alpha < - E_s(X_{\text{val}}^{i}) < \beta, ~ 
  X^{i}_{\text{val}}\in X_{\text{val}} \},
  \label{eq:xfb}
\end{align}
where $\beta$ is the upper threshold to select samples with low negative energy score (OOD samples) and $\alpha$ is a lower threshold to exclude samples that are very noisy and drastically OOD. The selected OOD samples ($X_\text{fb}$) along with their labels are then incorporated in the prompt $\mathcal{P}_{train}$ for generating train data in the next iteration. 

\subsection{SLM Training}
In each iteration, the generated train data ${X}^t_{train}$ is used to train and update the SLM. Since there might be noisy data labels inaccurately generated by the LLM, we use symmetric cross-entropy loss \cite{sce} for training the SLM in our work to address the potential noises in the samples generated by LLM. The symmetric cross-entropy loss consists of two terms including reverse cross-entropy and cross-entropy as follows:  
\vspace{-7pt}
\begin{align}
    \label{eq:sce}
    \mathcal{L}_{sce}=\frac{-1}{N}\sum_{i=1}^{N}\big(\lambda \sum_{k=1}^{K} \hat{y}_{k} \log(y_{k}) 
    \\ \nonumber
    + {\sigma} \sum_{k=1}^{K} {y}_{i,k} \log(\hat{y}_{i,k})\big)
    \vspace{-5pt}
\end{align}
where $\hat{y}$ and ${y}$ are respectively the predicted and ground-truth labels (generated as text). 
$K$ is the vocabulary size of the tokenizer of $\mathcal{M}_S$, $N$ is the total number of samples, and $\lambda$ and $\sigma$ are hyperparameters that adjust the weight of reverse cross entropy and cross-entropy. We set $\lambda$=1.0 and $\sigma$=0.1, which are obtained experimentally.

\section{Experiments}

In this section, the performance of \emph{GOLD} is analyzed and compared with previous works over different classification and sequence-to-sequence tasks. First, we describe the experimental settings including the datasets (tasks), model architectures, prompts, and baselines. Following that, the experimental results are quantitatively and qualitatively discussed and compared with the baselines. An extensive set of ablations over \emph{GOLD} components, size of LLM and SLM, and the number of generated samples is also provided.


\begin{table*}[!h]
\footnotesize
    \begin{center}
        \begin{tabular}{lcc|l|p{0.7cm}p{0.8cm}p{0.7cm}p{0.8cm}p{0.5cm}p{1cm}|p{0.5cm}}
          \toprule
          \textbf{Model}&\textbf{Model Size}&\textbf{Data}&\textbf{Method}&\textbf{ANLI}&\textbf{MNLI}&\textbf{QNLI}&\textbf{WNLI}&\textbf{RTE}&\textbf{MRPC}&\textbf{Ave.}\\ \midrule
          T5-base&220M&Full&Fine-Tuned&43.1&86.6&93.7&78.8&80.1&87.5 &{78.3}\\ 
          &&&Pre-Trained&29.0&56.6&88.3&52.1&68.5&75.0 &{61.6}\\ \midrule
           LLaMA2&7B&-&Few-shot & 36.0& 41.5&55.3&53.5& 62.4&65.4&52.3\\ 
           \midrule
           \multirow{3}{*}{LLaMA2 $\rightarrow$ T5-base}&\multirow{3}{*}{220M}&\multirow{3}{*}{3K}&P2Model&  34.4& 59.5& 62.2 &56.3&58.8&75.0&57.7\\
           &&&ZeroGen&34.6&56.1&88.5&54.9&62.1&84.3&63.4\\
          &&&ProGen&34.3&55.1&85.9&{57.7}&66.0&80.3&63.2\\
          &&&\textbf{\emph{GOLD}}&\textbf{35.7}& \textbf{62.5}& \textbf{91.7}&\textbf{57.7}&\textbf{69.6}&\textbf{85.0}&{\textbf{67.1}} \\
          \bottomrule
        \end{tabular}
        \vspace{-3pt}
        \caption{Comparison results on classification tasks in terms of accuracy. {We use T5-base (220M) as SLM and LLaMa2 (7B) as LLM. \textit{Pre-Trained}: simultaneously pre-trained on a variety of datasets including all the downstream ones (except ANLI and WNLI). \textit{Fine-Tuned}: specifically fine-tuned over the downstream dataset. \textit{Few-shot}: inference with the LLM given a few data samples.}}
        \label{tab:classification}
    \end{center}
\end{table*}

\begin{table*}[!h]
\begin{center}
\small
\begin{tabular}{lcc|l|cc|ccc}
      \toprule
      \textbf{Model}& \textbf{Model Size}&\textbf{Data}&\textbf{Method}&\textbf{SQUAD}&\textbf{Adv-QA}&\textbf{SVAMP}&\textbf{NL4OPT}\\ 
      \textbf{}&&&&\textcolor{gray}{(EM\%)}&\textcolor{gray}{(EM\%) }&\textcolor{gray}{(Acc\%)}&\textcolor{gray}{(ROUGE-L)}\\ \midrule
      T5-base&220M&Full&Fine-Tuned&77.3&38.5&53.7&89.2 \\ 
      &&&Pre-Trained&74.7&25.3&-&- \\ \midrule
      LLaMA2&7B&-&Few-shot &{15.2*}&{14.6*}&27.3&54.4\\ \midrule
     
      \multirow{5}{*}{LLaMA2 $\rightarrow$ T5-base}&\multirow{5}{*}{220M}&\multirow{5}{*}{3K}&ZeroGen &69.4 &21.3&20.0&67.9\\
      &&&ProGen &68.1&  20.5&23.7&68.5\\
      &&&P2Model&74.4&25.0&\textbf{26.0}&71.5\\
      &&&\textbf{\emph{GOLD}}&\textbf{75.2}&\textbf{25.5}&25.3&\textbf{72.8}   \\ \bottomrule
\end{tabular}
\vspace{-3pt}
\caption{Comparison results on seq-to-seq tasks. {\textit{EM}: Exact Match. \textbf{Note}: pre-trained T5-base has not seen SVAMP \& NL4OPT tasks (nor any similar dataset). {*: obtained with the general prompts in Table \ref{tab:prompts}. EM of 54.7 \& 25.1 is respectively achieved for SQUAD and Adv-QA in case of using specific prompts in \cite{llama2}}}.}
\label{tab:seq2seq}
 \end{center}
 \vspace{-3pt}
\end{table*}

\subsection{Settings}
\label{ssec:settings}
\textbf{Tasks.} We use 6 different classification tasks including ANLI, MNLI, QNLI, WNLI, RTE, and MRPC \citep{glue}, where accuracy is used as the evaluation metric. We further use 3 sequence-to-sequence tasks including SQUAD \cite{squad}, Adversarial QA (Adv-QA) \cite{advqa}, and SVAMP \cite{svamp}. SQUAD and Adv-QA are question and answer {(QA)} datasets while SVAMP is a math word problem. We respectively use Exact Match {(EM)} and accuracy as the evaluation metrics for QA and SVAMP datasets. In order to further evaluate the performance of \emph{GOLD} on less explored novel tasks, we perform experiments on recently released dataset NL4OPT \cite{nl4opt}. Input and output of NL4OPT dataset are respectively description of an optimization problem and its corresponding optimization formulation. We use ROUGE-L \citep{lin2004rouge} as the evaluation metric for this task. {The validation sets from the above-mentioned datasets are used for the numerical analysis of all the methods in this paper.}

\textbf{LLM and SLM.} We respectively use LLaMA2-7B \citep{llama2} and {pre-trained} T5-base (220M) \citep{t5} as the LLM and SLM for all our experiments. As T5 is a sequence-to-sequence model, we define all of the tasks including the classification ones as sequence-to-sequence tasks. Doing so, our framework is flexible to be applied to various NLP tasks. {Details of the datasets used for pre-training the T5 model used in this paper are provided in \cite{t5}.}

\textbf{Initial Prompts.} For all the experiments in this paper, we use a fixed, general prompt including a task definition and 3 samples of real data. Table \ref{tab:prompts} shows an example prompt for the ANLI task used by \emph{GOLD} and the other methods. We use the same prompt for different methods for fair comparison. The prompts for other tasks are given in Appendix \ref{sec:qual_results_sup}.

\textbf{Baselines.} We compare our results with three prior arts including ZeroGen \cite{zerogen}, ProGen \cite{progen}, and Prompt2Model \cite{prompt2model}. Although ZeroGen and ProGen are initially designed for classification tasks, we modify their implementation to be further applicable to sequence-to-sequence tasks. {We also compare our results with the pre-trained SLM, fine-tuned SLM, and few-shot performance of the specified LLM (i.e., LLaMA2-7B). The pre-trained SLM is indeed the model that has been simultaneously pre-trained on a variety of datasets including all the downstream ones (except ANLI and WNLI). On the other hand, the fine-tuned version is the model specifically fine-tuned over a downstream dataset. Moreover, the few-shot results of the LLM is achieved by doing the inference with the LLM given a few data samples.} {Unless otherwise mentioned, for the experiments corresponding to all the tasks and methods in this paper, we generated 3K samples using the LLM in 375 iterations.}

\begin{table*}[!h]
\footnotesize
    \centering
    \begin{tabular}{p{1.05cm}|p{0.09cm}p{13.5cm}} 
    \toprule
            \textbf{Method}&\textbf{ID}&\textbf{Random Samples}\\ \midrule
            \multirow{2}{*}{ZeroGen}&1& \textcolor{olive}{\textbf{Sentence1:} The new smartphone from Apple has a cutting-edge AI assistant that can learn and adapt to the user's preferences. , \textbf{Sentence2:} The AI assistant from Samsung is more advanced and can perform complex tasks with greater ease. , \textbf{Label:} entailment} \\
            &2& \textcolor{olive}{\textbf{Sentence1:}  The new smartphone features a cutting-edge AI-powered camera that can automatically detect and enhance low-light photos., \textbf{Sentence2:} The latest smartphone has a revolutionary camera that can take high-quality photos in any lighting conditions., \textbf{Label:} neutral} \\
            \midrule
            
            \multirow{2}{*}{ProGen}&1& \textcolor{teal}{\textbf{Sentence1:} The new smartphone has a powerful processor that enables it to run complex apps with ease., \textbf{Sentence2:} The new smartphone has a sleek design that makes it easy to hold and use., \textbf{Label:} neutral} \\
            &2& \textcolor{teal}{\textbf{Sentence1:} The new smartphone from Apple is revolutionizing the way people communicate. , \textbf{Sentence2:}  The new smartphone from Samsung is a total failure and will not last more than a year., \textbf{Label:}contradiction} \\
            \midrule
            \multirow{2}{*}{P2Model}&1& \textcolor{brown}{\textbf{Sentence1:} The company's new marketing strategy aims to increase sales by targeting a younger demographic., \textbf{Sentence2:} The company will see an increase in sales as a result of the new marketing strategy., \textbf{Label:} entailment} \\
            &2& \textcolor{brown}{\textbf{Sentence1:} The new AI model can generate realistic images of objects and scenes based on text descriptions. , \textbf{Sentence2:}  The model can create images of objects and scenes that are indistinguishable from real-world examples., \textbf{Label:} entailment} \\
             \midrule
            \multirow{2}{*}{\emph{\textbf{GOLD}}}&1& \textcolor{darkblue}{\textbf{Sentence1:}  The 2022 Winter Olympics are scheduled to take place in Beijing, China from February 4 to 20, 2022., \textbf{Sentence2:} The 2022 Winter Olympics are expected to be a showcase of the latest technology and innovation in the sporting world. , \textbf{Label:} neutral} \\
            &2& \textcolor{darkblue}{\textbf{Sentence1:} The new electric car model is environmentally friendly and reduces carbon emissions., \textbf{Sentence2:} The new electric car model is powered by fossil fuels and emits more carbon than a traditional gasoline-powered car., \textbf{Label:} contradiction} 
        \\\bottomrule
    \end{tabular}
    \vspace{-3pt}
    \caption{Two data samples generated by ZeroGen, ProGen, and \emph{GOLD} randomly sampled from the generated dataset {for ANLI task.} \emph{GOLD} generates more diverse samples compared to baseline methods.  
    \vspace{4pt}
    }
    \label{tab:qualitative}
\end{table*}

\subsection{Results}

\textbf{Quantitative Results.} 
Table \ref{tab:classification} shows the results of our method on  the classification tasks. The results are compared with the prior arts, the LLM, and the pre-trained/fine-tuned SLMs. 
{It is shown that our method (with an average accuracy of 67.1\%) improves the performance of the pre-trained SLM (with an average accuracy of 61.6\%) over all the six tasks. Our method demonstrates a 4\% improvement over ZeroGen and ProGen, and also 10\% improvement over P2Model.} Compared to the LLM's few-shot with an average accuracy of 52.3\%, we substantially obtain better results with 14\% margin.
{Despite the LLM's poor performance over the validation sets of the tasks, the data generated by the LLM is sufficiently accurate to facilitate knowledge transfer to the SLM. To study this, we used GPT-4 \citep{gpt4} to generate the ground truth labels of the generated data, and then evaluated the LLM's accuracy. To this end, a significant accuracy of 80.3\% and 72.1\% was respectively observed over the data generated for QNLI and RTE.}

 \setlength{\tabcolsep}{5pt} 
\begin{table*}[!h]
\begin{center}
\footnotesize
\begin{tabular}{l|l|cccccccc|c}
      \toprule
      \textbf{Method}&\textbf{Model}&\textbf{ANLI}&\textbf{MNLI}&\textbf{QNLI}&\textbf{WNLI}&\textbf{RTE}&\textbf{MRPC}&\textbf{SVAMP}&\textbf{NL4OPT}&\textbf{Ave.}\\ \midrule
      {Pre-Trained}&T5-small (60M) &28.7&47.4&86.3&43.6&52.7&69.3&-&-&54.7\\ 
      &T5-base (220M) &29.0&56.6&88.3&52.1&68.5&75.0&-&-&61.6\\ 
      &T5-large (770M) &32.6&58.9&90.1&45.0&72.5&72.3&-&-&61.9\\ \midrule
      {\emph{\textbf{GOLD}}}&LLaMA2 $\rightarrow$T5-small &32.6&56.1&83.5&46.4&63.5&75.9&13.3&70.4&55.2\\ 
      &LLaMA2 $\rightarrow$T5-base  &{35.7}&62.5&91.7&\textbf{56.3}&69.6&\textbf{85.0}&25.3&\textbf{72.8}&62.4\\ 
      &LLaMA2 $\rightarrow$T5-large&\textbf{35.7}&\textbf{63.3}&\textbf{93.3}&52.1&\textbf{74.3}&83.1&\textbf{29.7}&72.5&\textbf{63.0}\\ 
    \bottomrule
\end{tabular}
\vspace{-5pt}
\caption{{The performance of \emph{GOLD} with different SLM sizes. The bigger the model, the better the performance (except WNLI and MRPC that follow the same trend as in the pre-trained SLMs).}}
\label{tab:ablation_slm}
\end{center}
\end{table*}

The comparison results for the sequence-to-sequence tasks are given in Table \ref{tab:seq2seq}, where \emph{GOLD} is better than prior arts, few-shot results of the LLM, and pre-trained SLM on SQUAD and Adv-QA datasets. Note that the evaluation metric of SQUAD and Adv-QA datasets is Exact Match (EM) that requires the output to be exactly similar to the labels. {Moreover, as mentioned in Section \ref{ssec:settings}, we used a consistent prompt format in all of our experiments including the LLM's (i.e., LLaMA2) few-shot results for simplicity and generality. Therefore, LLaMA2 provides lower EM accuracy (i.e., 15.2 and 14.6 for SQUAD and Adv-QA) compared to the results reported in their original paper \cite{llama2}. However, if we use the specific prompts from LLaMA2's paper, the results are improved to 54.7 and 25.1 for SQUAD and Adv-QA, respectively.
}

Note that the SVAMP dataset was not included in the train set of the pre-trained SLM. However, the LLM may have been exposed to similar datasets during its training phase. \emph{GOLD} (with an accuracy of 25.3\%) is the second best on SVAMP dataset after P2Model (with an accuracy of 26.0\%). 



Table \ref{tab:seq2seq} also presents the results on NL4OPT, a recently released dataset. The SLM has not encountered this dataset or a similar task before. Likewise, the LLM probably has not been exposed to this dataset or this specific task, although it might have encountered optimization problems from other sources. We aim to explore the potential of \emph{GOLD} in preparing SLMs for new tasks that the LLM has not directly encountered before. Our proposed method outperforms prior works by obtaining a ROUGE-L of 72.8\%. More results on other seq-to-seq tasks are given in Appendix \ref{sec:medical_sup}.


As expected, the best performance is achieved by the SLMs specifically fine-tuned on the train set of each downstream dataset in a supervised way, i.e., the upper-bound results in this work.

\textbf{Qualitative Results.} Table \ref{tab:qualitative} shows two samples that were randomly selected from the datasets generated by ZeroGen, ProGen, P2Model, and \emph{GOLD}. The samples generated by ProGen and ZeroGen are closely related, while those produced by P2Model and \emph{GOLD} exhibit greater diversity. More results are given in Appendix \ref{sec:qual_results_sup}.

\subsection{Ablations}

\textbf{SLM Size.} Table \ref{tab:ablation_slm} shows the 
experimental results with different SLM models including T5-small (60M), T5-base (220M), and T5-large (770M). 
{In general, as the SLM size increases, better results are obtained by both pre-trained and distilled versions. However, there are two exceptions including WNLI and MRPC for which T5-base outperforms T5-large. This is due to the overfitting of the larger model on the small datasets of WNLI and MRPC.}

\setlength{\tabcolsep}{3.4pt} 
\begin{table}[!t]
\small
\begin{center}
\begin{tabular}{c|ccc|cccc}
      \toprule

    &\textbf{KD}&\textbf{FB}&\textbf{SCE}&\textbf{MNLI}&\textbf{QNLI}&\textbf{SVAMP}&\textbf{RTE}\\ \midrule
    V0 &&&&56.6&88.3&-&68.5 \\ \midrule
    V1 &\checkmark&&&55.4&90.6&20.0&67.1\\
    V2&\checkmark&\checkmark& &59.5&91.4&23.3&68.6\\
    V3&\checkmark&&\checkmark&59.1&91.0&17.3&63.1\\
     \textbf{\emph{GOLD}}&\checkmark&\checkmark&\checkmark &\textbf{62.5}&\textbf{91.7}&\textbf{25.3}&\textbf{69.6} \\ \bottomrule

\end{tabular}
\caption{Ablation on the main components of \emph{GOLD}. \textit{V0}: pre-trained SLMs. \textit{KD}:  knowledge distillation by vanilla data generation. \textit{FB}: the feedback function in the course of data generation. \textit{SCE}: symmetric cross entropy used for training the distilled SLM.}
\label{tab:ablation}
 \end{center}
\end{table}

\begin{figure*}[!h]
    \centering
    \includegraphics[scale=0.4]{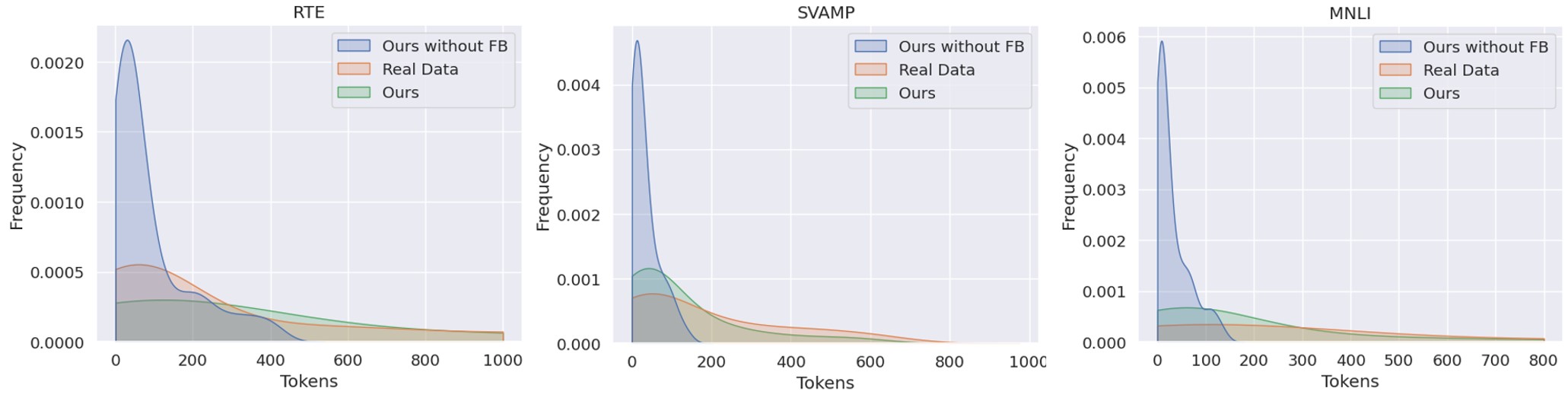}
    \vspace{-5pt}
    \caption{
    The distribution of generated data by our method with and without OOD-based feedback. 
    }
    \label{fig:distribution}
\end{figure*}

\textbf{Components of \emph{GOLD}.} Table \ref{tab:ablation} summarizes the ablation on the main components of \emph{GOLD}. Ablating the feedback function (i.e., V3 in the table) decreases the accuracy on RTE and MNLI datasets by 6\% and 3\%, respectively. Similarly, ablating the noise-robust SCE loss (i.e., V2) decreases the accuracy by about 3\% on RTE and MNLI datasets. This ablation study shows that having noise-robust loss is critical when fine-tuning SLMs using synthetic generated data. Moreover, the feedback function is effective to boost the performance. 

\textbf{Distribution of Generated Data.} Figure \ref{fig:distribution} illustrates the distribution of the data generated by \emph{GOLD} with and without OOD-based feedback compared with that of real data. 
{The distribution plots for example classification and sequence-to-sequence tasks are shown. As noted in the plots, the data generated using our proposed OOD-based feedback mechanism provides a more long-tailed distribution (i.e., including low-probability events) that is closer to the real data distribution. In contrast, vanilla data generation results in only high-likelihood data, while disappearing low probability tokens. More plots are given in Appendix \ref{sec:dist_plots_sup}.}

\setlength{\tabcolsep}{2.0pt} 
\begin{table}[!h]
\begin{center}
\footnotesize
\begin{tabular}{l|l|ccc}
      \toprule
      \textbf{Method}&\textbf{Model} &\textbf{MNLI}&\textbf{WNLI}&\textbf{RTE}\\ \midrule
      Few-shot&LLaMA2-7B &41.5&53.5&62.4\\ 
      &LLaMA2-70B&55.6&63.4&74.7\\ \midrule 
      \emph{\textbf{GOLD}}&LLaMA2-7B $\rightarrow$ T5-base&60.0&57.7&69.3\\ 
      &LLaMA2-70B$\rightarrow$ T5-base&\textbf{62.4}&\textbf{65.7}&\textbf{71.1}\\ 
    \bottomrule
\end{tabular}
\vspace{-5pt}
\caption{Ablation on LLM size with 1K samples. 
}
\label{tab:ablation_llm}
\end{center}
\end{table}

\textbf{LLM Size.} 
Table \ref{tab:ablation_llm} presents an ablation study conducted on the size of the LLM. We carried out the experiments with only 1K samples using two different versions of LLM: LLaMA-70B and LLaMA-7B. The results indicate that the few-shot performance of the 70B LLM surpasses that of the 7B one. Furthermore, when the SLM is distilled from the 70B LLM, it demonstrates superior performance compared to the SLM distilled from the 7B LLM. This suggests that our framework can potentially enhance the results when a larger LLM is employed. However, it is worth noting that it becomes increasingly challenging to exceed the few-shot results of larger LLMs. For instance, as shown in Table \ref{tab:ablation_llm}, the distilled SLM consistently outperforms the few-shot performance of LLM when the 7B version is used. However, this trend does not hold when the 70B LLM is utilized.

\textbf{Dataset Size.}
Table \ref{tab:datasetsize} presents the performance of our method when the volume of generated data varies from 0 to 5K samples. 
{As observed, increasing the number of samples (that are mostly OOD) does not necessarily improve the overall performance of \emph{GOLD}. 
However, this is only the case for the tasks on which the SLM has been pre-trained using real data. In other words, our pre-trained SLMs have already seen enough high likelihood samples, which are supposed to be the majority compared to the OOD samples. Thus, there is no need for big amount of OOD samples and a small, but robust set of samples is sufficient to improve the generalizability of the distilled SLM. On the other hand, for the SVAMP that the pre-trained SLM has not encountered before, increasing the volume of data consistently enhances the accuracy.}

\setlength{\tabcolsep}{5pt}
\begin{table}[!h]
\begin{center}
\small
\begin{tabular}{l|cccc}
      \toprule
      \textbf{\#Samples}&\textbf{MNLI}&\textbf{SVAMP}&\textbf{QNLI}&\textbf{RTE}\\ \midrule
      0&56.6&0&88.3&68.5\\
      1000&60.0&17.3&91.5&69.3\\ 
      2000&\textbf{63.4}&22.7&90.4&69.3\\ 
      3000&62.5&25.3&\textbf{91.7}&\textbf{69.6}\\ 
      4000&62.5&25.3&90.9&68.9\\ 
      5000&62.9&\textbf{28.3}&89.4&69.3\\ 
      \bottomrule
\end{tabular}
\vspace{-6pt}
\caption{Ablation on the number of generated samples.}
\label{tab:datasetsize}
\end{center}
\end{table}

\vspace{-4pt}
\subsection{Hyperparameters Selection}
\textbf{$\alpha$ and $\beta$ in Eq. \ref{eq:xfb}.}
We utilize a lower and an upper threshold, denoted as $\alpha$ and $\beta$, respectively, in Eq. \ref{eq:xfb} for the selection of OOD samples. For each batch of data, we set these thresholds to select samples that fall within the range of 50\% to 80\% of the lowest negative energy scores. {As shown in \cite{curseofrecursion}, most of the samples generated by LLMs are high-likelihood samples. Therefore, specifying the upper threshold as 50\% guarantees that all of the excluded samples are high likelihood. Indeed, extensive hyperparameter tuning on each downstream dataset improves the results. However, for simplicity and generality, we used a fixed set of hyperparameters on all tasks.}

\textbf{$\sigma$ and $\lambda$ in Eq. \ref{eq:sce}.}
{As we do not use any validation set to select the hyperparameters, the upper and lower thresholds in Eq. 8 were chosen based on \cite{sce}, where $\sigma$=0.1 and $\lambda$=1 in most of the experiments. \cite{sce} shows that in symmetric cross-entropy loss, large $\sigma$ (e.g., between 0.5 and 1.0) tends to cause more overfitting, while small $\sigma$ (e.g., between 0.01 and 0.1) can help ease the overfitting of CE. We also experimented on different values of $\sigma$, while fixing $\lambda$ to 1 over the ANLI dataset. As shown in Table \ref{tab:hp_eq8},  $\sigma$=0.1 provides the best performance. Overall, the value $\sigma$ does not have a significant effect on our method's performance and we still outperform prior arts with different values.}

\setlength{\tabcolsep}{5pt}
\begin{table}[!h]
\begin{center}
\small
\begin{tabular}{l|cccc}
      \toprule
      \textbf{$\sigma$}& 0.01 & 0.1 & 1\\ \midrule
      \textbf{Accuracy} & 0.354 & \textbf{0.357} & 0.350\\
      \bottomrule
\end{tabular}
\vspace{-6pt}
\caption{Analysis of $\sigma$ and $\lambda$ values in Eq. \ref{eq:sce}.}
\label{tab:hp_eq8}
\end{center}
\end{table}


\vspace{-4pt}
\subsection{Running Time Analysis}
{Table \ref{tab:runtime} presents the running time, in terms of seconds per sample, of \emph{GOLD} and prior works. The analysis was done on 4 NVIDIA Titan Xp GPUs. As seen, ProGen exhibits the highest running time, which is due to using an expensive influence function to measure the quality of generated samples. On the other hand, ZeroGen and Prompt2Model, which lack any feedback and evaluation function and do not append feedback samples to the prompt, are slightly faster than \emph{GOLD}.}

{Note that the time cost of \emph{GOLD} and the previous works linearly increases with the number of samples. Consequently, the total time required to generate 3K samples for each task is approximately 4 hours for ZeroGen, around 4.5 hours for both \emph{GOLD} and Prompt2Model and about 6 hours for ProGen. It should also be noted that we only append 2-3 OOD samples from the previous iteration to generate a new batch of data, which only adds an overhead of 0.4 seconds per sample (20min for 3k samples).}

\setlength{\tabcolsep}{5pt}
\begin{table}[!h]
\begin{center}
\small
\begin{tabular}{l|cccc}
      \toprule
      \textbf{Method}&{ZeroGen}&{ProGen}&{P2Model}&{\emph{GOLD}}\\ \midrule
      \textbf{Time (s)} & 4.9 & 7.3 & 5.3 & 5.4\\
      \bottomrule
\end{tabular}
\vspace{-6pt}
\caption{Running time analysis.}
\label{tab:runtime}
\end{center}
\end{table}

\vspace{-5pt}
\subsection{Lexical Diversity}
Table \ref{tab:diversity} shows the lexical diversity of our generated data compared to prior arts. Lexical diversity is defined as the number of unique words divided by the total number of words in the dataset. {P2Model} generates the most diverse samples of data due to the high temperature value, and \emph{GOLD} is the second best in terms of the diversity of data. P2Model increases the temperature of the LLM during data generation to enhance diversity. However, we argue that relying on temperature can potentially compromise the accuracy of the generated samples. {To this end, we utilized GPT-4 to evaluate the accuracy of the data generated by the LLM for P2Model with a high temperature, where an accuracy of 38.6\% and 59\% was respectively obtained on the QNLI and RTE. For our method, where the LLM temperature is fixed to the default value of 1, we achieved an accuracy of 80.3\% and 72.1\% for the same tasks.}

\begin{table}[!h]
\footnotesize
\begin{center}
\begin{tabular}{l|cccc}
      \toprule
      &\multicolumn{4}{c}{\textbf{Lexical Diversity}} \\ 
    \textbf{Method}&MNLI&QNLI&MRPC&SQUAD\\ \midrule
     ZeroGen&1.5&3.7&2.7&1.3\\
      ProGen&1.5&3.7&2.8&1.9\\
      P2Model&\textbf{7.6}&10.2&\textbf{7.0}&\textbf{10.0}\\
      \textbf{\emph{GOLD}}&\underline{6.4}&\textbf{11.5}&\underline{6.6}&\underline{3.1} \\ \bottomrule

\end{tabular}
\vspace{-3pt}
\caption{Lexical diversity of the generated data by \emph{GOLD} and prior works. {P2Model uses high temperature for more diversity.} 
}
\label{tab:diversity}
 \end{center}
\end{table}

\vspace{-5pt}
\section{Conclusion} 
In this paper, we proposed a task-agnostic framework for data generation and KD from LLMs to SLMs that is applicable to any NLP task, including new tasks. In order to improve the generalizability of the distilled SLMs, we introduced an iterative OOD-empowered feedback mechanism. An energy-based OOD evaluation approach was also proposed to handle noisy data generated by LLMs. With an extensive set of experiments, we showed that our method achieved state-of-the-art results on a variety of classification and sequence-to-sequence downstream tasks in NLP. Future direction of this work includes applying the same technique for other data modalities such as images.

\clearpage
\section{Limitation} 
Despite its promising results, \emph{GOLD} does have certain limitations that warrant further investigation. One of the primary challenges we encountered was identifying the failure modes of the SLM and subsequently guiding the data generation procedure. This approach is susceptible to generating samples that are drastically different, which can negatively impact the performance of the distilled SLM. To mitigate this issue, we implemented an upper-bound threshold for selecting OOD samples. However, our observations indicate that some data points in the feedback loop still deviate from the correct format of real data.

Future work should focus on incorporating a data valuation method. This would enable the automatic evaluation of selected samples before they are fed back into the LLM. Such approach could significantly enhance the reliability and accuracy of the data generation process, leading to improved performance of the distilled models.

In this study, we demonstrated the potential of preparing SLMs for NL4OPT, a novel task for which the LLM has not been specifically trained. This approach opens up new avenues for leveraging language models in diverse applications. However, the performance of our proposed method on such innovative tasks requires further investigation. Future research should focus on evaluating the effectiveness of \emph{GOLD} across a broader range of user-defined tasks.



\bibliography{anthology,custom}

\newpage
\appendix

\section{Appendix Introduction}
In this appendix, we provide 1) more quantitative results on text generation tasks, 2) more distribution plots of the generated data, 3) the specific prompts utilized for each task, and 4) further qualitative results.

To ensure the reproducibility of our experiments, we have also shared the relevant code with detailed instructions, {which is available}
\href{https://developer.huaweicloud.com/develop/aigallery/notebook/detail?id=9d770d1f-3758-4d0f-99d4-3346abbe1546}
{here\footnote{\url{https://developer.huaweicloud.com/develop/aigallery/notebook/detail?id=9d770d1f-3758-4d0f-99d4-3346abbe1546}}}. 

\section{Results on Text Generation Tasks}
\label{sec:medical_sup}
{
We have conducted experiments on text-to-text tasks, namely NL4OPT and SVAMP in the main body of the paper. In this section, we conduct another set of experiments on a recently released, challenging text generation task, known as Medical-Dialogue-to-Note \cite{abacha2023empirical}. The objective of this task is to generate a note that a doctor would take during a patient visit, given a dialogue between the patient and the doctor. This dataset has been recently released, and we assume that both the LLM and SLM have not been exposed to this dataset. As shown in Table \ref{tab:medical}, the SLM (i.e., T5-base) distilled using \emph{GOLD} achieves a ROUGE-L score of 0.198, which is twice better than the pre-trained T5 with 0.101 and is comparable with the LLM (i.e., LLaMA2-7B) few-shot results with 0.218.}

\begin{figure}[h!]
    \centering
    \includegraphics[scale=0.7]{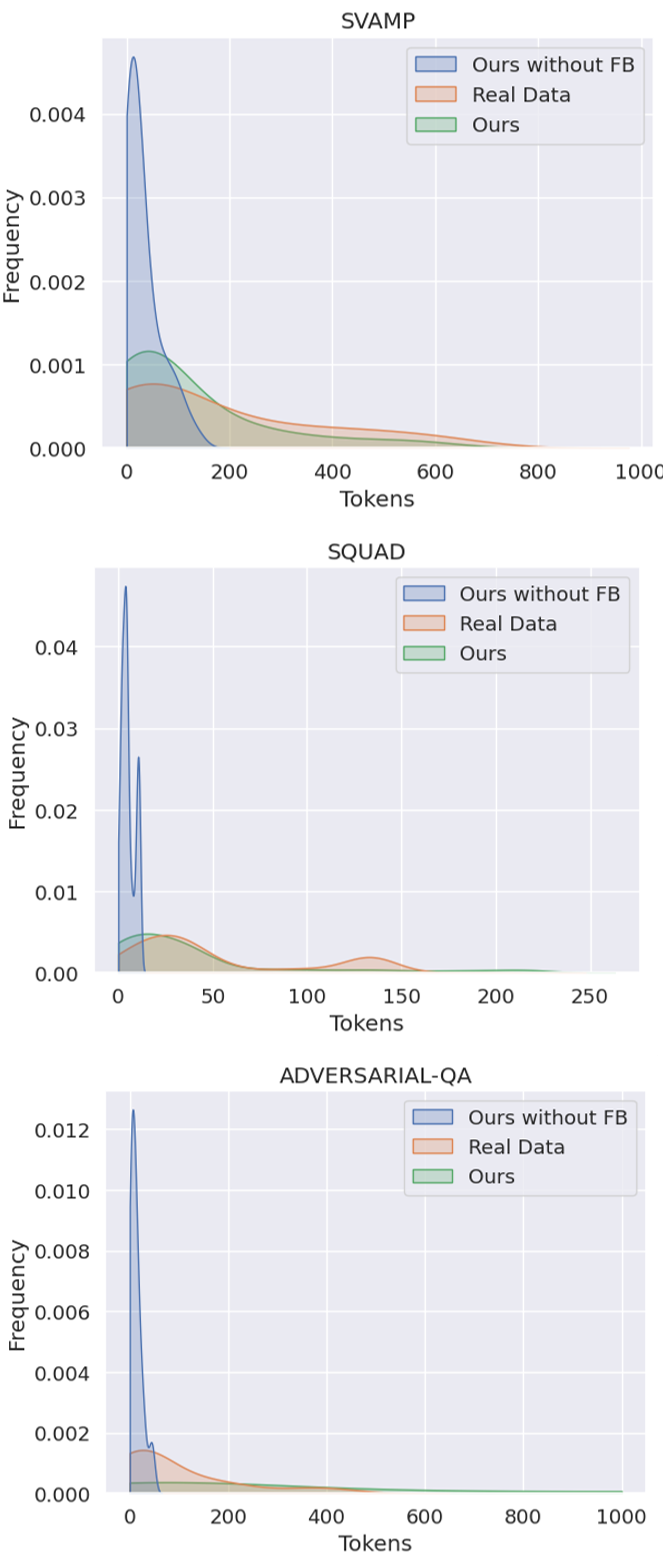}
    \caption{The distribution of generated data by our method with and without OOD feedback.}
    \label{fig:distribution4}
\end{figure}

\setlength{\tabcolsep}{5pt}
\begin{table}[H]
\begin{center}
\small
\begin{tabular}{l|cccc}
      \toprule
      &PT-SLM & FT-SLM & LLM & {\emph{GOLD}}\\ \midrule
      \textbf{ROUGE-L} &0.101 &0.329 & 0.218 & 0.198\\
      \bottomrule
\end{tabular}
\vspace{-6pt}
\caption{Results on Medical-Dialogue-to-Note text generation task. PT: pre-trained; FT: fine-tuned.}
\label{tab:medical}
\end{center}
\end{table}

\section{Generated Data Distribution}
\label{sec:dist_plots_sup}
Figure \ref{fig:distribution4} illustrates the distribution plots of the data generated by \emph{GOLD} with and without OOD-based feedback compared with that of real data. The distribution plots for sequence-to-sequence tasks including SVAMP, SQUAD, and Adv-QA are shown.

\section{Prompts and Qualitative Results}
\label{sec:qual_results_sup}
Tables \ref{tab:prompts3}, \ref{tab:prompts2}, and \ref{tab:prompts4} provide the detailed prompts that were used to generate data for each task. We provide three real samples from the train set of each task in the prompt. When we have OOD samples as the feedback, they will be added after the real samples. Also, Table \ref{tab:samples} provides samples of data generated for each of the tasks.  

\begin{table*}[!h]
\small
    \centering
    \begin{tabular}{p{1.5cm}p{1.5cm}|p{10cm}} 
    \toprule
        Data&Method &Prompt \\ \midrule
        \multirow{4}{*}{NL4OPT}& \emph{GOLD}, 
      
        ZeroGen, ProGen, P2Model&\textcolor{blue}{The above are examples of NL4OPT dataseta and below are three samples of 'NL4OPT' dataset. The samples have a 'problem' which is a linear optimization problem and a 'formulation' which is the formulation of the optimization problem.}

        \textcolor{olive}{1.
        problem: A hotel employs cleaners and receptionists. Cleaners earn \$500 per week and receptionists earn \$350 per week. The hotel requires a minimum of 100 workers of whom at least 20 must be receptionists. To keep the hotel clean and running smoothly, the number of receptionists should be at least a third of the number of cleaners. The hotel wants to keep the weekly wage bill below \$30000. Formulate a LP to minimize the wage bill. 
        formulation: Variables: x: sled dogs, y: trucks, Objective Function: maximize 100  x + 300  y ,Constraints:50  x + 100  y $\leq$ 1000, x  $\leq$  y }

        \textcolor{olive}{2. problem: An office supply company makes two types of printers: color printers and black and white printers. Different sections of the factory with different teams produce each printer. The color printer team can produce at most 20 color printers per day while the black and white printer team can produce at most 30 black and white printers per day. Both teams require use of the same paper tray installing machine and this machine can make at most 35 printers of either type each day. Color printers generate a profit of \$200 per printer while black and white printers generate a profit of \$70 per printer. How many of each printer should be made to maximize the company's profit? formulation: Variables: x: color printers, y: black and white printers, Objective Function: maximize 200  x + 70  y, Constraints:  x $\leq$ 20,  y $\leq$ 30,  x +  y $\leq$ 35}

        \textcolor{olive}{ 3. problem: An accounting firm has senior accountants earning \$3000 per week and junior accountants earning \$1000 per week. The contracts with companies to provide accounting services require at least 100 accountants, of whom at least 5 must be senior accountants. To make sure there is enough experience on the accounting team, the number of senior accountants should be at least a third of the number to junior accountants. The firm wants to keep the weekly wage bill below \$150000. Formulate an LP to minimize the wage bill. formulation: Variables: x: senior accountants, y: junior accountants, Objective Function:minimize 3000  x + 1000  y, Constraints:  x +  y $\geq$ 100,  x $\geq$ 51,  x  $\geq$ 0.33  y, 3000  x + 1000, y $\leq$ 150000.
        \textcolor{blue}{
        The above are samples of  NL4OPT data.  Think step by step and give me a novel sample of NL4OPT dataset.} 
        } 
          \\ \hline 
      
 \bottomrule
    \end{tabular}
    \caption{Sample prompts used for NL4OPT dataset. We use the same prompt for different methods for fair comaprison. }
    \label{tab:prompts3}
\end{table*}

\begin{table*}[!h]
\small
    \centering
    \begin{tabular}{p{1.5cm}p{1.5cm}|p{10cm}} 
    \toprule
        Data&Method &Prompt \\ \midrule
        \multirow{4}{*}{QNLI}& \emph{GOLD}, 
        ZeroGen, ProGen, P2Model&\textcolor{blue}{ Below are three samples of QNLI dataset. Samples include a Question and a Sentence. The label of sample is 1.'entailment' if the answer of the Question is in the Sentence and 2.'not\_entailment' if the answer of the Question is not in the Sentence. }
        \textcolor{olive}{ 1. Sentence: He must do this by collecting the multiple Tears of Light; once all the Tears of Light are collected for one area, he restores that area's Light Spirit. Question: What does Link have to gather in order to complete each area? Label: entailment}  
        \textcolor{olive}{2. Sentence:Prior to this time congressional parties were often relatively disorganized, so it was not always evident who functioned as the opposition floor leader. Question: Why was minority leader position created?. Label: entailment  }
        \textcolor{olive}{3. Sentence:This view is shared by other researchers who argue that the ancestors of the American Indians were the first to separate from the great Asian population in the Middle Paleolithic. Question:Who have studies of the mtDNA of Turkic-speaking peoples shown they're closest to genetically? Label: not\_entailment}  
        \textcolor{blue}{The above are three samples of QNLI data.  Think step by step and give me a novel sample of QNLI data with <not\_entailment> label. }  \\ \hline 
        \multirow{1}{*}{RTE-WNLI}& \emph{GOLD}, 
        ZeroGen, ProGen, P2Model&\textcolor{blue}{You are a helpful 'Assistant'. You only reply once as 'Assistant'. Do not pretend to be a 'User'. Below are three samples of RTE dataset. Samples include a 'Sentence1' and a 'Sentence2'. The label of sample is 'entailment' if the answer of the 'Sentence1' entails 'Sentence2' and 'not\_entailment' if 'Sentence1' does not entail 'Sentence2'.} 
        \textcolor{olive}{
         1. Sentence1: No Weapons of Mass Destruction Found in Iraq Yet. Sentence2: Weapons of Mass Destruction Found in Iraq. Label: not\_entailment}
         \textcolor{olive}{
         2. Sentence1: A place of sorrow, after Pope John Paul II died, became a place of celebration, as Roman Catholic faithful gathered in downtown Chicago to mark the installation of new Pope Benedict XVI. Sentence2: Pope Benedict XVI is the new leader of the Roman Catholic Church.  Label: entailment }
         \textcolor{olive}{
         3. Sentence1:Herceptin was already approved to treat the sickest breast cancer patients, and the company said, Monday, it will discuss with federal regulators the possibility of prescribing the drug for more breast cancer patients. Sentence2:Herceptin can be used to treat breast cancer. Label: entailment 
         \textcolor{blue}{
         The above are three samples of RTE data. Think step by step and give me a novel sample of RTE data with <entailment> label.}
         }  \\ \hline
        \multirow{2}{*}{MRPC}& \emph{GOLD}, 
        ZeroGen, ProGen, P2Model&\textcolor{blue}{Below are three samples of MRPC dataset. Samples include a 'Sentence 1' and a 'Sentence 2'. The label of sample is 'equivalent' if the 'Sentence 1' and the 'Sentence 2' are paraphrases of each other. The label is 'not\_equivalent' if 'Sentence 1' and 'Sentence 2' are not semantically equivalent.}
        \textcolor{olive}{1. Sentence 1: Amrozi accused his brother , whom he called  the witness , of deliberately distorting his evidence . Sentence 2: Referring to him as only the witness , Amrozi accused his brother of deliberately distorting his evidence . Label: equivalent }
        \textcolor{olive}{2. Sentence 1: Yucaipa owned Dominick 's before selling the chain to Safeway in 1998 for \$ 2.5 billion . Sentence 2: Yucaipa bought Dominick 's in 1995 for \$ 693 million and sold it to Safeway for \$ 1.8 billion in 1998 . Label: not\_equivalent}
        \textcolor{olive}{3. Sentence 1: Around 0335 GMT , Tab shares were up 19 cents , or 4.4 \% , at A \$ 4.56 , having earlier set a record high of A \$ 4.57 . Sentence 2: Tab shares jumped 20 cents , or 4.6 \% , to set a record closing high at A \$ 4.57 . Label: not\_equivalent}  
        \textcolor{blue}{The above are three samples of MRPC data. Think step by step and give me a novel sample of RTE data with <equivalent> label.}  
        \\ \hline
        \multirow{2}{*}{SVAMP}& \emph{GOLD}, 
        ZeroGen, ProGen, P2Model&\textcolor{blue}{
        Below are 3 examples of SVAMP dataset. Samples include a 'Body' which  explains a simple math problem and a 'Quesiton' which ask a question from the 'Body'.  }
        \textcolor{olive}{
        1. Body:There are 87 oranges and 290 bananas in Philip's collection. If the bananas are organized into 2 groups and oranges are organized into 93 groups. Question: How big is each group of bananas? Equation:( 290.0 / 2.0 )}  
        \textcolor{olive}{
        2.  Body: Marco and his dad went strawberry picking. Marco's dad's strawberries weighed 11 pounds. If together their strawberries weighed 30 pounds. Question: How much did Marco's strawberries weigh? Equation: ( 30.0 - 11.0 ) }
        \textcolor{olive}{ 
        3. Body: Edward spent \$ 6 to buy 2 books each book costing him the same amount of money. Now he has \$ 12. Question: How much did each book cost? Equation: ( 6.0 / 2.0 )  }
        \textcolor{blue}{
        The above are samples of  SVAMP data.  Think step by step and give me a novel sample of SVAMP dataset.}  \\
 \bottomrule
    \end{tabular}
    \caption{Sample prompts used for QNLI, RTE, WNLI, MRPC, and SVAMP dataset. We use the same prompt for different methods for fair comaprison. }
    \label{tab:prompts2}
\end{table*}

\begin{table*}[!h]
\small
    \centering
    \begin{tabular}{p{1.5cm}p{1.5cm}|p{10cm}} 
    \toprule
        Data&Method &Prompt \\ \midrule

        \multirow{4}{*}{SQUAD}& \emph{GOLD}, 
      
        ZeroGen, ProGen, P2Model&\textcolor{blue}{Below are samples of SQUAD data. It has a 'context' which is a paragraph from wikipedia, a 'question' from the paragraph and a short 'answer' for the question. 'answers' are directly from the 'context'.}

        \textcolor{olive}{1. context: Architecturally, the school has a Catholic character. Atop the Main Building's gold dome is a golden statue of the Virgin Mary. Immediately in front of the Main Building and facing it, is a copper statue of Christ with arms upraised with the legend 'Venite Ad Me Omnes'. Next to the Main Building is the Basilica of the Sacred Heart. Immediately behind the basilica is the Grotto, a Marian place of prayer and reflection. It is a replica of the grotto at Lourdes, France where the Virgin Mary reputedly appeared to Saint Bernadette Soubirous in 1858. At the end of the main drive (and in a direct line that connects through 3 statues and the Gold Dome), is a simple, modern stone statue of Mary. question:To whom did the Virgin Mary allegedly appear in 1858 in Lourdes France? answer:Saint Bernadette Soubirous }

        \textcolor{olive}{2. context: As at most other universities, Notre Dame's students run a number of news media outlets. The nine student-run outlets include three newspapers, both a radio and television station, and several magazines and journals. Begun as a one-page journal in September 1876, the Scholastic magazine is issued twice monthly and claims to be the oldest continuous collegiate publication in the United States. The other magazine, The Juggler, is released twice a year and focuses on student literature and artwork. The Dome yearbook is published annually. The newspapers have varying publication interests, with The Observer published daily and mainly reporting university and other news, and staffed by students from both Notre Dame and Saint Mary's College. Unlike Scholastic and The Dome, The Observer is an independent publication and does not have a faculty advisor or any editorial oversight from the University. In 1987, when some students believed that The Observer began to show a conservative bias, a liberal newspaper, Common Sense was published. Likewise, in 2003, when other students believed that the paper showed a liberal bias, the conservative paper Irish Rover went into production. Neither paper is published as often as The Observer; however, all three are distributed to all students. Finally, in Spring 2008 an undergraduate journal for political science research, Beyond Politics, made its debut. question: In what year did the student paper Common Sense begin publication at Notre Dame? answer: 1987 }

        \textcolor{olive}{3. context: The College of Engineering was established in 1920, however, early courses in civil and mechanical engineering were a part of the College of Science since the 1870s. Today the college, housed in the Fitzpatrick, Cushing, and Stinson-Remick Halls of Engineering, includes five departments of study – aerospace and mechanical engineering, chemical and biomolecular engineering, civil engineering and geological sciences, computer science and engineering, and electrical engineering – with eight B.S. degrees offered. Additionally, the college offers five-year dual degree programs with the Colleges of Arts and Letters and of Business awarding additional B.A. and Master of Business Administration (MBA) degrees, respectively. question: How many departments are within the Stinson-Remick Hall of Engineering? answer: fiven }

        \textcolor{blue}{The above are samples of  SQUAD  data.  Think step by step and give me a novel sample of SQUAD dataset.} 
          \\ \hline

 \bottomrule
    \end{tabular}
    \caption{Sample prompts used for SQUAD and Adv-QA datasets. We use the same prompt for different methods for fair comaprison. }
    \label{tab:prompts4}
\end{table*}


\begin{table*}[!h]
\small
    \centering
    \begin{tabular}{p{1.2cm}|p{0.1cm}p{13cm}} 
    \toprule
            \textbf{Method}&\textbf{ID}&\textbf{Random Samples}\\ 
            \midrule
            \multirow{2}{*}{QNLI}&1&{\textbf{Sentence:} "The Titanic was designed to move at speeds of up to 21 knots and could accommodate more than 3,500 passengers and crew.", 
            
            \textbf{Question:} "What was the maximum speed of the Titanic?", 
            
            \textbf{Label:} "entailment"} \\
            &2&{\textbf{Sentence:} "The new manager is very friendly and always willing to help.",
            
            \textbf{Question:} "How many years of experience does the new manager have in the field?",
            
            \textbf{Label:} "not\_entailment"} \\
            \midrule
            \multirow{2}{*}{RTE}&1&{\textbf{Sentence1:} "The use of digital technologies in agriculture has the potential to revolutionize food production and reduce the environmental impact of farming.", 
            
            \textbf{Sentence2:} "The development of digital agriculture will lead to increased food availability and better nutrition for the world's population.", 
            
            \textbf{Label:} "entailment"} \\
            &2&{\textbf{Sentence1:} "There are several reasons behind the growing trend of veganism, including the health benefits of a plant-based diet, the ethical concerns surrounding animal welfare, and the environmental impact of livestock production.", 
            
            \textbf{Sentence2:} "The consumption of dairy products leads to better bone health.",
            
            \textbf{Label:} "not\_entailment"} \\
            \midrule
            \multirow{2}{*}{MRPC}&1&{\textbf{Sentence1:} "The new AI technology has the potential to revolutionize healthcare.", 
            
            \textbf{Sentence2:} "The latest advancement in artificial intelligence is expected to transform the medical field.", 
            
            \textbf{Label:} "equivalent"} \\
            &2&{\textbf{Sentence1:} "The company will launch a new product line next quarter.", 
            
            \textbf{Sentence2:} "The company will lay off 10\% of its workforce next month.", 
            
            \textbf{Label:} "not\_equivalent"} \\
            \midrule
            \multirow{2}{*}{SVAMP}&1&{\textbf{Body}: "A car travels 480 kilometers in 8 hours.", 
            
            \textbf{Question}: "How far does the car travel per hour?", 
            
            \textbf{Equation}: " ( 480.0 / 8.0 ) ", 
            
            \textbf{Answer}: "60"}\\
            &2&{\textbf{Body}: "If Sally has 35 apples and she gives 10 to her friend.", 
            
            \textbf{Question}: "How many apples does Sally have left?", 
            
            \textbf{Equation}: " ( 35.0 - 10.0 )  ",
            
            \textbf{Answer}: "25"}\\
            \midrule
            \multirow{2}{*}{NL4OPT}&1&{\textbf{Problem}:  "A company produces two products, A and B, with different production rates. The company has a total of 10 machines, and each machine can produce either product. The production rate of product A is 2 machines per day, while the production rate of product B is 3 machines per day. The company wants to maximize the total daily production of both products."
            
            \textbf{Formulation}:
            
            Variables: 
            
            x: number of machines producing product A
            
            y: number of machines producing product B 
    
            Objective Function:
            
            maximize $2.00 * x + 3.00 * y$
            
            Constraints:
            
            $x + y \leq 10.00$
            
            $x \geq 0$
            
            $y \geq 0$
            } \\
           &2&{\textbf{Problem}:  "A company produces two types of smartphones, A and B. Smartphone A has a battery life of 24 hours and a profit margin of \$50 per unit, while smartphone B has a battery life of 12 hours and a profit margin of \$30 per unit. The company has a total budget of \$100000 for producing these smartphones. The company wants to determine the number of units of each smartphone to produce to maximize the profit.",
           
           \textbf{Formulation}:
            
            Variables:
            
            x:  number of units of smartphone A
            
            y:  number of units of smartphone B
            
            Objective Function:
            
            maximize $(50.00 * x + 30.00 * y)$
            
            Constraints:
            
            $x + y\leq 100000.00$} 
        \\\bottomrule
    \end{tabular}
    \caption{Two data samples generated by \emph{GOLD} randomly sampled from the generated dataset for different datasets. 
    }
    \label{tab:samples}
\end{table*}

\end{document}